\documentclass[lettersize,journal]{IEEEtran}
\usepackage{amsmath,amsfonts}
\usepackage{algpseudocode}
\usepackage[ruled,vlined]{algorithm2e}
\usepackage{stfloats}
\usepackage{array}
\usepackage{textcomp}
\usepackage{url}
\usepackage{verbatim}
\usepackage{lipsum} 
\usepackage{multicol}
\usepackage{hyperref}
\usepackage{orcidlink}
\usepackage{graphicx}
\usepackage{caption}
\usepackage{subcaption}
\usepackage{cuted}
\usepackage{amssymb}
\usepackage{amsmath}
\usepackage{wasysym}
\usepackage{diagbox}
\newcommand{\full}{\CIRCLE}
\newcommand{\part}{\LEFTcircle}
\newcommand{\none}{\Circle}
\usepackage{tikz}
\usepackage{pifont}    

\usetikzlibrary{trees, arrows.meta, positioning}

\tikzset{
  basic/.style  = {draw, text width=8em, align=center, rounded corners, minimum height=1.5em},
  root/.style   = {basic, font=\bfseries, fill=blue!10},
  level1/.style = {basic, fill=blue!10},
  level2/.style = {basic, fill=blue!5},
  connect/.style = {->, thick}
}
\usepackage{multirow} 
\usepackage{float} 
\usepackage{cite}

\usepackage{mwe}      
\usepackage{booktabs}  
\begin{document}

\title{RAGAM-AI: A Hybrid Reinforcement Learning based Agentic approach powered by Generative AI for Edge Resource Management }
\author{Shreya K. Chari\orcidlink{0009-0005-9667-2016},~\IEEEmembership{Student Member,~IEEE,}
        John S. Vardakas\orcidlink{0000-0003-1656-0507},~\IEEEmembership{Senior Member,~IEEE,}
        Kostas Ramantas\orcidlink{0000-0002-1304-784X},
        Christos Verikoukis\orcidlink{0000-0001-8774-1052},~\IEEEmembership{Senior Member,~IEEE}
        
        \IEEEcompsocitemizethanks{\IEEEcompsocthanksitem S. Chari is with Iquadrat Informatica S.L., Barcelona, Spain, and with the Universitat Politècnica de Catalunya, Barcelona, Spain, \IEEEcompsocthanksitem J. Vardakas is with the Department of Informatics of the University of Western Macedonia, Greece, and with Iquadrat Informatica S.L., Barcelona, Spain, \IEEEcompsocthanksitem K. Ramantas is with Iquadrat Informatica S.L., Barcelona, Spain,
\IEEEcompsocthanksitem C. Verikoukis is with the University of Patras, Patras, Greece, and with the Industrial Systems Institute (ISI) $|$ Athena Research Center, Patras, Greece}
 }
\maketitle
\begin{abstract}
Resource management for distributed edge networks hosting different services has been handled efficiently using Reinforcement learning (RL). Reward generalization in complex workflows is still an open issue within the RL realm. Therefore, this work proposes RAGAM-AI, a hybrid framework that combines the capabilities of RL algorithms and Generative-AI (Gen-AI) for powering an Agentic AI workflow that optimizes resource allocation at the Multi-access Edge Computing (MEC) units. This framework further resolves the reward generalization issue by utilizing Large Language Model (LLM) for feedback. A Multi-Agent Soft-Actor Critic (MASAC) using Math$\Sigma$tral as a reward oracle is implemented for managing resource allocations at the MEC. RAGAM-AI interprets system state information and evaluates multi-objective trade-offs guiding the resource orchestration loop over distributed edge server agents. The complexity of the framework is further brought down by incorporating surrogate reward model. The overall performance is evaluated against the single agent SAC, REINFORCE and random policies for benchmarking. On using RAGAM-AI, lower allocation deviations for CPU, memory and bandwidth resources are reported while minimizing the latency and data rate costs. A scalable AI as a feedback (AIAF) solution optimizing network resources with significant performance improvement is thus presented.               
\end{abstract}  


\section{Introduction}
\IEEEPARstart{T}HE limitations presented by a centralized cloud architecture discouraged localized processing of tasks increasing latency and congestion within networks. In such a scenario, a MEC at the edge of the network emerged as a solution for bringing computation and devices closer enabling task offloading in resource-constrained conditions \cite{Peng-etal}. MEC is further in control of network resources such as compute power, bandwidth and memory. However, efficient resource management within MEC yet remains as a critical and challenging problem.\\
\indent The long-established systems fail to consider non-stationary network traffic distributions and the coupling effect between different resources \cite{Fateneh-etal}. Heterogeneous nodes now cater to diverse use-cases that require low-latencies and faster compute capacities. Many of these issues are resolved at the edge network with the help of MEC servers. Long transmission time, energy expenditures and data breaches are alleviated by partially or fully offloaded tasks to the edge \cite{Dong-etal}. On the other hand, edge caching relieves delays caused from the backhaul pressure significantly by fetching frequently accessed data \cite{Zhang-etal}. Fault tolerance and system resiliency can be improved by having task distribution and decision-making closer to the network rather than a centralized cloud \cite{Patricia-etal}. There is an enhanced context-awareness as information with regards to local radio link signals and network traffic load bursts become more available \cite{mustafa-etal}.    \\
\indent There are numerous use-cases that benefit from putting edge computing into operation. In Vehicle-to-everything (V2X), smart transportation systems' edge computing has enabled low latency and high reliability safety services \cite{seo-etal}. Edge networks can support real-time fault-detection and load balancing in smart grids alleviating network outage issues \cite{dande-etal}. For running inferences, analytics and predictive maintenance, Industrial IoT (IIoT) often rely on edge powered solutions \cite{F-zhang-etal}. Edge caching is often used for Extended Reality (XR) technologies for providing rendering support \cite{duru-etal}. Digital twins require localized optimization for data-ingestion and for processing sensor outputs relying on edge support \cite{chen-etal}.  \\
\indent However, a wide range of open challenges continues to limit practical deployment of these use-cases. Various network aspects such as compute, storage, communication and energy are optimized in isolation due to independently functioning Radio Access Network (RAN) and MEC orchestrators \cite{jiang-etal}. Network handover and service migration are not yet completely service-aware, causing disruption in providing quality services to the end user \cite{zhao-etal}. The privacy and security concerns at the edge in distributed settings are also issues of concern in order to make the systems more fault resilient \cite{deng-etal}. Limitations on the standards and the available 5G architectures make the cross-domain interactions complex leading to unsatisfied Service Level Agreements (SLA) \cite{sasian-etal}.\\
\indent RL has emerged as a promising solution to bridge some of the gaps mentioned above. Resource allocation is now possible subject to dynamic network conditions using RL's learning abilities even under fluctuating workloads \cite{jiang-etal}. Multi-objective optimization involving both the communication and compute requirements can be achieved using RL \cite{tan-etal}. Decentralized distributed multi-agent RL system enables localized learning at each node contributing to global objectives in dense edge deployments \cite{chari-etal}. RL-based approaches have also showcased reduction in overall latencies and energy expenditures with better resource utilization \cite{khan-etal}. \\
\indent The advantages of using RL algorithms at the edge got further enhanced with the availability of Generative modeling. RL algorithms often are constrained by partial observability, limited training data and lack of explainability. Generative models can assist RL algorithms by synthesizing unforeseen scenarios and help in  generalization \cite{xu-etal}. RL algorithms involve interactive learning with the environment often making it sample inefficient and expensive \cite{qin-etal}. Gen-AI aided intervention can alleviate this problem by optimizing the search spaces and reducing some of the interactions. \\ 
\indent In recent times, there's an increasing interest in autonomous frameworks that can manage and execute sophisticated decisions entirely on its own. These systems give rise to concepts like Agentic-AI systems and Mixture-of-Experts (MoE) which breakdown complex tasks into independent sub-tasks at both the training and evaluation level \cite{lo-etal}. Further, the performance enhancement is achieved using valuable feedback through methods like Reinforcement Learning from Human Feedback (RLHF) and Reinforcement Learning from AI Feedback (RLAIF) \cite{hao-etal,bai-etal}. In RLHF, direct human assessment is included as a feedback creating scalability and expense issues. On the other hand, RLAIF involves AI assistants for supervising the algorithm behaviour.   \\
\indent Keeping in mind all the discussion above, it can be established that resource management at the network edge is still a fundamental challenge as the networks becomes more heterogeneous with diverse performance constraints. RL algorithms provide a promising relief for adaptive decision-making while still coping with reward bias and explainability challenges. The studies \cite{thomas-etal,laidlaw-etal} confirm that imperfect or rigid reward formulations can indeed affect the learning capabilities of the RL policies. These issues are especially relevant in edge resource orchestration, where static reward shaping may inadequately capture dynamic trade-offs among latency, utilization, fairness, and demand variability. Recent advances in Gen-AI when combined with existing RL solutions can provide the required performance enhancement improving the sample efficiency and the semantic reasoning for generalization.  \\
\indent Therefore, in this work, a hybrid framework utilizing the learning capabilities of a distributed Multi-Agent Deep RL (MADRL) at the MEC is integrated with the feedback from GenAI-based LLM model extending the preliminary results in \cite{chari-genai-etal}. The framework incorporates multiple interacting RL agents operating across edge nodes driven by feedback from the LLM. The policy updates are coordinated through shared environmental objectives and reward signals. The framework therefore showcases core agentic properties of autonomy, feedback inclusion, task distribution and adaptive behavior. The framework thus achieved is interpreted as an agentic-AI system successful in eliminating the reward bias and in improving the overall algorithm performance measured in terms of efficient allocations, latency and data rate. The salient features of the framework implemented are as follows:     
\begin{enumerate}
    \item A RL-based SAC multi-agent system using AI as feedback by utilizing the Math$\Sigma$tral driven reward oracle is wrapped in a hybrid Agentic AI framework. 
    \item The developed RAGAM-AI algorithm focuses on managing edge network resources: CPU, memory and bandwidth efficiently. 
    \item Resource deviation, latency cost and data rate cost is used for performance evaluation. The results show that RAGAM-AI outperforms the considered benchmarking policies. The average values for improvements observed are summarized in Table \ref{tab_sectionIntro}.
    \item The problem of reward bias is alleviated by the inclusion of LLMs in the reward oracle which is further optimized by training a supervised reward model.
\end{enumerate}

\begin{table}[t]
\centering
\caption{Summary of results}
\label{tab_sectionIntro}
\resizebox{\columnwidth}{!}{
\begin{tabular}{|c|c|c|c|c|}
\hline
\centering
\diagbox{\textbf{Metrics}$\downarrow$}{\textbf{Policy}} & \textit{RAGAM-AI} & \textit{SAC} & \textit{REINFORCE} & \textit{Random} \\
\hline
\textit{CPU dev.} & 23.65\%  & 28.61\% & 32.53\% & 44.93\%\\
\hline
\textit{Memory dev.} & 20.36\%  & 24.35\% & 24.68\% & 46.37\%\\
\hline
\textit{Bandwidth dev.} & 20.40\% & 25.27\% & 26.5\% & 46.92\%\\
\hline
\textit{Latency cost} & 6.70e-3 & 6.72e-3 &6.724e-3 & 8.12e-3\\
\hline
\textit{Data rate cost} & 1.151e+8 & 1.153e+8 & 1.154e+8 & 1.228e+8\\
\hline
\end{tabular}
}
\end{table}

The remainder of this paper is organized as follows. Section II discusses the related works focusing on different approaches for resource management at network edge. Section III gives an overview of the system design used for the proposed framework in this paper. Section IV lays out the details of the framework and algorithm thus designed with relevant background required for understanding. The performance results and comparisons are further analyzed in Section V. Towards the end, Section VI provides conclusion and future research directions with respect to this paper.

\begin{table*}[htbp]
\centering
\caption{Summary of related work}
\resizebox{\textwidth}{!}{
\begin{tabular}{c|c|c|c|c|c|c}
\hline
\hline
   \textbf{Reference} & \textbf{Application} & \multicolumn{4}{c|}{\textbf{Highlights}}  & \textbf{Metric evaluated} \\
\cline{3-6}

& &  \textbf{Resource allocation} & \textbf{Task offloading} & \textbf{Expandability} & \textbf{Heterogeneity} & \\
\hline
\cite{jiang-etal} & Network slicing & \full  & \none & \part & \part & Processing ratio \\
\hline
\cite{yu-etal} & IoT devices & \full  & \full & \none & \none & Task completion time, energy consumption \\
\hline
\cite{yibo-zhang-etal} & UAV, D2D & \part  & \part & \none & \part & Task offloading ratio, computational latency \\
\hline
\cite{chu-etal} & Distributed systems & \full  & \full & \part & \none & System gain, delay, energy\\
\hline
\cite{zeng-etal} & Ultra-dense low latency networks & \part  & \none & \part & \none & Resource utilization, load reduction, energy\\
\hline
\cite{Elkawkagy-etal} & Latency sensitive users & \full  & \full & \none & \part & Resource utilization, task completion time \& offloading proportion \\
\hline
\cite{wang-etal} & Cloud-native orchestration & \full  & \none & \full & \full & Resource allocation, latency \\
\hline
\cite{W-zhang-etal} & Mobile user offloading & \full  & \full & \part & \part & Workload balance, system cost \\
\hline
\cite{dai-etal} & Digital twin & \part  & \full & \part & \none & System cost, latency cost, energy cost \\
\hline
\cite{yang-etal} & IoT & \full  & \full & \part & \none & Execution utility \& time \\
\hline
\cite{liu-etal} & Vehicular edge computing & \full  & \full & \part & \none & Inference time, backlogged computations \\
\hline
\cite{nguyen-etal} & Blockchain-mining & \part & \full & \part & \none & Offloading utility, Network bandwidth cost, system utility \\
\hline
\cite{cheng-etal} & Augmented reality & \part & \none & \part & \none &  Utility, frame rate, accuracy, latency\\
\hline
\cite{huang-etal} & SAGINs & \full & \part & \part & \none &  Energy, latency\\
\hline
\cite{x-zheng-etal} & Cloud-edge collaboration & \none & \part & \full & \none &  Task performance, QoE\\
\hline
\cite{z-liu-etal} & AI-generated content & \part & \none & \part & \part &  Hit ratio, objective function performance\\
\hline
\cite{tang-etal} & Consumer electronic devices & \part & \none & \part & \none &  Cost of renting, makespan, energy consumption\\
\hline
\cite{du-etal} & Metaverse & \none & \part & \part & \none &  Reward, task performance\\
\hline
\hline
\end{tabular}
}
{\footnotesize \full: Fully supported, \part: Partially supported, \none: Not supported}

\label{rw1}

\end{table*}

\begin{table*}[htbp]
\caption{Limitations \& improvement from related work}
\resizebox{\textwidth}{!}{
\begin{tabular}{m{2cm}|m{6cm}|m{6cm}}  
\hline
\hline
 \begin{center} \textbf{Reference} \end{center} & \begin{center}  \textbf{Identified limitations} \end{center} &  \begin{center} \textbf{Improvement using RAGAM-AI}\end{center}\\
\hline
\centering\cite{yu-etal}-\cite{Elkawkagy-etal} & \begin{itemize}
    \item Scalability concerns 
    \item Static or semi-static network assumptions
    \item Security and privacy Concerns
    \item High computational complexity
    \item Parameter sensitivity
\end{itemize}  
 & \begin{itemize}
    \item Model-free optimization
    \item Improved handling of high-dimensional spaces
    \item Automatic policy learning
\end{itemize}\\
\hline
\centering\cite{jiang-etal}, \cite{wang-etal}-\cite{huang-etal}     & \begin{itemize}
    \item Information exchange during centralized training
    \item Training instabilities from hierarchical layers
    \item Hybrid-action spaces
    \item Synchronization issues
    \item High algorithmic complexity
    \item Deployment and Integration Challenges
\end{itemize} & \begin{itemize}
    \item Sequential decision-making
    \item Adaptation to changing environment
    \item Elimination of labelled datasets
    \item Focus on long-term objectives
\end{itemize}\\
\hline
\centering\cite{x-zheng-etal}-\cite{du-etal} & \begin{itemize}
    \item Model maintenance and updates
    \item Computational overhead at the edge
    \item Stationarity considerations
\end{itemize} & \begin{itemize}
    \item Distributed learning 
    \item Improved reasoning 
    \item Handling complex state information
    \item Avoiding reward bias
    \item Including AI as feedback
\end{itemize}   \\
\hline
\hline
\end{tabular}
}
\label{rw2}
\end{table*}
\section{Related work}
In this section, the related literature prior to this work is reviewed and summarized. This section attempts to showcase the evolution of resource management at the network edge from the initial traditional optimization approaches to the current amalgamation with AI-guided management. Further, the strengths, highlights and limitations of these works are discussed with a summary in Table \ref{rw1} and \ref{rw2}. The section ends with a careful consideration of how the research idea in this work fills the gaps in the literature by proposing significant improvements. 
\subsection{Edge resource management}
Attempts have been made using traditional optimization techniques to manage and orchestrate the edge network resources in various works. The authors of \cite{yu-etal}, propose a multi-objective evolutionary optimization framework to resolve the task offloading and resource allocation issues at the MEC servers. The developed decomposition-based evolutionary algorithm thus developed minimizes average task completion time and energy consumption of mobile devices. A MEC architecture catering to Unmanned Aerial Vehicles (UAVs) and Device-to-Device (D2D) communication for efficient resource utilization is proposed in \cite{yibo-zhang-etal}. A joint optimization problem integrating trajectory design, resource allocation, and task offloading decisions is formulated minimizing computation latency.\\
\indent In \cite{chu-etal}, user Quality of Experience (QoE) in federated MEC systems is jointly optimized using service caching, computational resource allocation, and task offloading. A novel two-stage algorithmic framework based on approximation and decomposition techniques framed as a Mixed-Integer Non Linear Programming (MINLP) with multi-layer constraints. A resource-efficient joint service caching and workload scheduling framework for ultra-dense edge networks is proposed in \cite{zeng-etal}. The developed CoShare algorithm reduces the workload on the cloud by utilizing Lyapunov technique for optimization.  
The research work in \cite{Elkawkagy-etal} uses genetic algorithm for joint optimization of offloading ratios, wireless bandwidth allocation, and computational resource distribution. Simulation results demonstrates reduction in task completion time while maintaining performance.
\subsection{AI-driven resource optimization using RL}
Over time, many of the open challenges were mitigated by making use of RL advancements. A two-level slicing model is introduced in \cite{jiang-etal} for dynamic network-slicing and resource allocation. The network slicing is resolved with a QMIX-based RL algorithm while the resource allocation is resolved using a Proximal Policy Optimization (PPO) based algorithm. In \cite{wang-etal}, a cloud-native architecture explores resource allocation efficiency and network resource utilization. The architecture is validated by testing it on a testbed to demonstrate the improvements. The authors of \cite{W-zhang-etal} use hierarchical RL to tackle joint caching and resource allocation issues in MEC systems. In recent times digital twins have gained popularity leading to high compute requirements at the edge. In \cite{dai-etal}, a federated deep RL scheme is designed to optimize task offloading, configurations of passive reflecting links, and computing resources from a two-layer digital twin edge network. A  Knowledge-Defined Edge Computing Networks (KDECN) architecture along with a RL-based Long-term Computation Offloading and computation Resource Allocation (L-CORA) algorithm is presented in \cite{yang-etal}. This approach intends to manage large-scale networks ensuring long-term quality of computing service for user devices. \\
\indent In the context of Vehicular Edge Computing (VEC), Multi-Agent Diffusion-based Deep RL (MAD2RL) is proposed in \cite{liu-etal} to jointly optimize Deep Neural Net (DNN) partitioning, task offloading, and resource allocation. In addition to that, a Lyapunov optimization technique is initially used to decouple the long-term problem into a per-slot deterministic problem. A novel solution combining blockchain-mining and edge system improving the overall system utility is presented in \cite{nguyen-etal}. The multi-objective formulation thus developed jointly  optimizes offloading decision, channel selection, transmit power allocation, and computational resource allocation. In \cite{cheng-etal}, low-latency video communication ensuring video quality in Mobile Augmented Reality (MAR) applications is explored. A Gated Recurrent Unit-Augmented Twin Delayed Deep Deterministic Policy Gradient (GTD3) is used to resolve the challenges from the need for dynamic computational resources and limited availability of network bandwidth. The authors of \cite{huang-etal} propose a novel MADRL algorithm in the MEC infrastructure within Space-Air-Ground Integrated Networks (SAGINs). The ground user tasks are modelled as Directed Acyclic Graphs (DAGs) while reducing energy consumption and latency in MEC.

\subsection{GenAI augmented performance enhancement}
Introduction of GenAI applications in RL-based systems has brought tremendous shift in how intelligent systems learn, adapt, and make decisions. For instance, in \cite{x-zheng-etal}, an innovative framework called EdgeNetLLM is developed for efficiently adapting LLM models for the tasks of adaptive bitrate streaming (ABR) and cluster job scheduling (CJS) in mobile networks. Additionally, it can also adapt to specific tasks and inputs minimizing the overall costs and latency. In \cite{z-liu-etal}, a Two-Timescale Deep RL (T2DRL) algorithm is decomposed as a caching subproblem on a long-timescale and a resource allocation subproblem on a short-timescale. The algorithm is designed for resource provisioning in edge-enabled AIGC service for reducing latency. Similarly, the authors of \cite{tang-etal} propose a lightweight LLM to augment RL for independent task scheduling minimizing makespan, operational costs, and energy consumption. To address the issue of GenAI models being resource-intense, \cite{du-etal} present a AI-Generated Optimal Decision (AGOD) algorithm for optimally select service providers. The AGOD algorithm utilizes Deep Diffusion Soft Actor-Critic (D2SAC) for increasing the efficiency of the selection made.   
\subsection{Research gap \& contributions}
This work improves the state-of-the-art literature by emphasizing on the need for performance enhancement of existing RL algorithms by taking advantage of the advancements and availability of pre-trained LLMs. RAGAM-AI is a model-free multi-agent approach that can function under dynamic network connections. The RL policy within RAGAM-AI can be conveniently scaled to multiple servers tackling training instabilities and growing state space. Further, the centralized training and decentralized execution within the algorithm enables distributed learning. RL algorithms are often entirely driven by the reward signals, thus a poorly designed signal either introduces learning bias or amplifies the ground truth. Therefore, RAGAM-AI uses a reward oracle that combines a traditional mathematically backed reward function alongside reward signals received from mindful prompting to the Math$\Sigma$tral model to minimize the reward generalization issue. Further, a small Deep Neural Network (DNN) is trained to mimic the reward function for lightweight architecture. The crucial contributions are listed below:
\begin{enumerate}
    \item Decentralized training mechanism by making use of MASAC algorithm with off-policy learning enabling scalability and sample efficiency. 
    \item Improved decision-making via hybrid Agentic AI formulation using the capabilities of interactive strategy updates through RL policies and context-based reasoning feedbacks from LLMs. 
    \item Optimized resource allocations minimizing the deviations from the resource demand while providing enhanced latency and data rate.  
\end{enumerate}
\section{System Design}
\begin{figure}[!t]
     \centering
        \includegraphics[ width=0.9\linewidth]{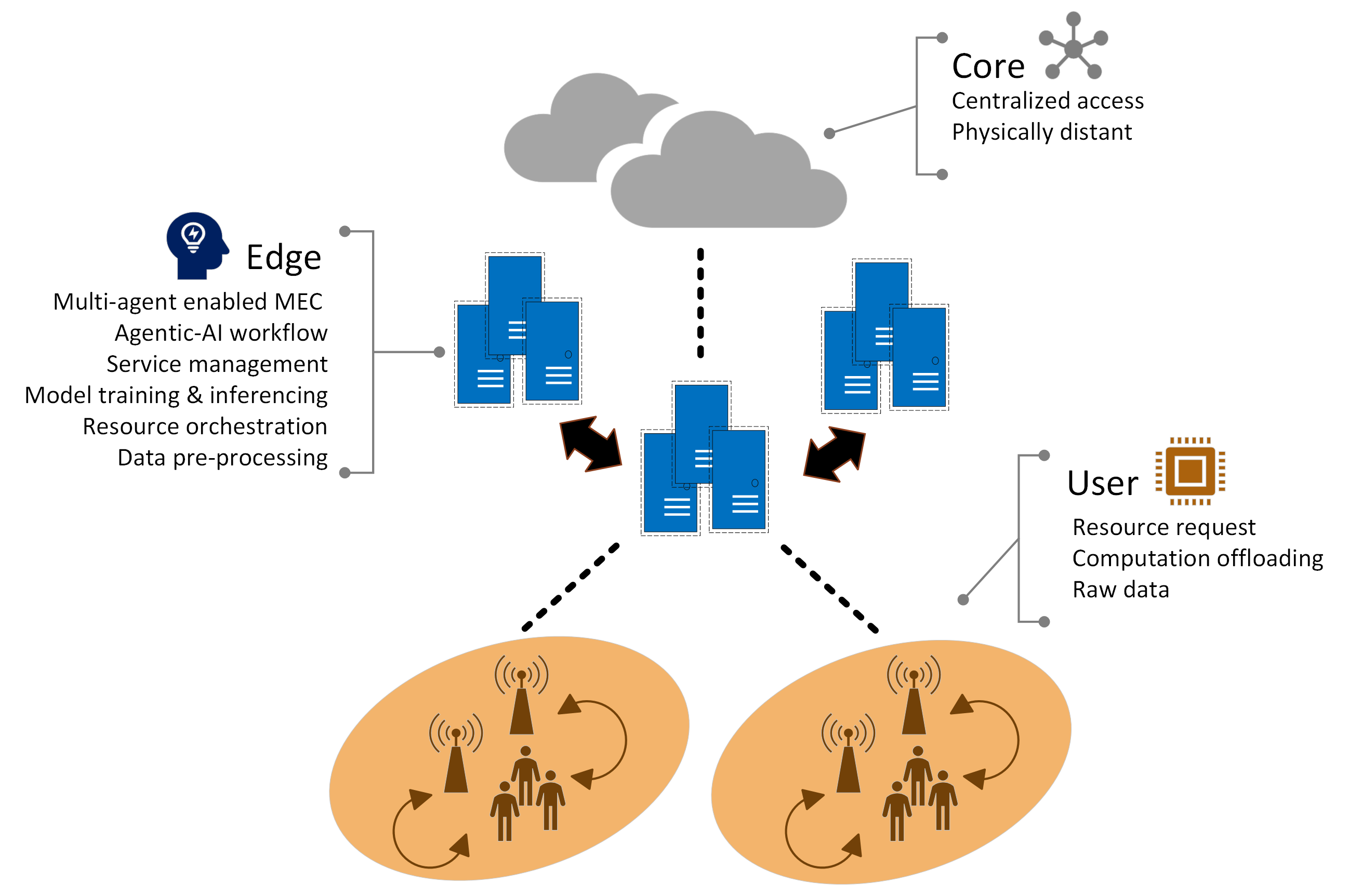}
       \caption{System design}
       \label{OC}
\end{figure}
In this section, the system model outline is provided, followed by the formal definition of the RAGAM-AI algorithm. The symbols used throughout this work are summarized in Table \ref{tab3}.  Additionally, Fig.\ref{OC} paints an overall visual how the system design looks with the most important components and their functionalities. The system focuses mainly on the resource requests received at the edge servers and the RL agents placed at each of the considered servers. The user requests are translated as edge-level resource utilization metrics, specifically CPU, memory, and bandwidth consumption through network services. Admission control is beyond the scope of this work, and it is assumed that all requests are admitted for convenience. 
\subsection{Problem definition}
The resource request intent from the users is mapped to the MEC servers $n_i$ as compute $c$, memory $m$ and bandwidth $b$ demands. At each time instance $t$ the request received can be visualized as in (\ref{2.1}). The requests are associated with network slices which are then admitted and orchestrated by at the edge. 
\begin{equation}
\begin{bmatrix}
(c, m, b)_{t}^{n_1} & (c, m, b)_{t}^{n_2} & \cdots & (c, m, b)_{t}^{N} \\
\vdots & \vdots & \ddots & \vdots \\
(c, m, b)_{T}^{n_1} & \cdots & \cdots & (c, m, b)_{T}^{N}
\end{bmatrix}
\label{2.1}
\end{equation}

\noindent The agents at each MEC server are constantly calculating the latency $L$ and data rate $D$ using Shannon equation in (\ref{2.21}) and (\ref{2.22}) to evaluate the agents decision making abilities \cite{shannon}. The $SNR$ is calculated using channel gains and the noise power $NP=-174+10\log_{10}B_{f}-NF$ where $NF$ is the noise figure. Further, $B$ denotes the effective bandwidth, defined as the product of the allocated bandwidth and the fixed channel bandwidth $B_{f}$. Standard values for the gains were used as mentioned in Table \ref{tableSM} \cite{standard}.
\begin{equation}
SNR=P_{TX}-PL+G_{TX}-NP
\label{2.21}
\end{equation}
\begin{equation}
D= B \,\log_{2}\bigl(1 + SNR\bigr)
\label{2.22}
\end{equation}
The proposed resource management problem is formalized as a minimization problem subject to constraints hereafter. It is assumed that the resources $a=(c_a,m_a,b_a)$ is assigned for each demand $d=(c_d,m_d,b_d)$. At a given time $t$, the difference between the allocated and demanded resources should be minimal provided the server capacities are not exceeded. Constraints \textit{C1} and \textit{C2} monitor the allocations by keeping them under the maximum capacity $C$ and more than the demand $d$. The measured latency $L$ and data rate $D$ thresholds are controlled by \textit{C3} and \textit{C4}. Finally, in \textit{C5}, the allocation of requests to the MEC servers is managed. All the incoming resource requests $\mathbb{N}$ with CPU, memory or bandwidth requirements are roughly divided as $x_{ij}$ between the available $\mathbb{M}$ edge servers. 
\begin{equation}
\begin{aligned}
&\min_{}\lvert a_t - d_t \rvert \ \\
\text{s.t. }\\
& \text{\textit{C1:}}\hspace{0.1cm} a_r \le C_r \quad 
 \text{\textit{C2:}}\hspace{0.1cm} a_r \ge d_r \quad \forall r \in \mathcal{R} \\
& \text{\textit{C3:}}\hspace{0.1cm} L \le \mathcal{L} \hspace{0.6cm}\text{\textit{C4:}}\hspace{0.1cm} D \ge \mathcal{D} \quad \\
& \text{\textit{C5:}}\left|\sum_i x_{ij}-\frac{\mathbb{N}}{\mathbb{M}}\right|\le 1 \qquad \forall j
\end{aligned}
\end{equation}

\begin{table}[t]
\centering
\caption{Summary of symbols }
\label{tab3}
\renewcommand{\arraystretch}{1.3}
\setlength{\tabcolsep}{8pt}
\begin{tabular}{ll}
\toprule
\textbf{Notation} & \textbf{Description} \\
\midrule
$n_i$ & MEC Servers \\
$c$ & CPU request\\
$m$ & Memory request\\
$b$ & Bandwidth request\\
$L$ & Latency\\
$\mathcal{L}$ & Latency limit\\
$D$ & Data rate\\
$\mathcal{D}$ & Data rate limit\\
$B$ & Effective bandwidth\\
$P_{tx}$ & Transmit power\\
$PL$ & Path loss\\
$G_{tx}$ & Antenna gain \\
$NF$ & Noise figure \\
$a$ & Resource allocations \\
$d$ & Resource demand \\
$x$ & Server allocation \\
$\mathbb{M}$ & Total available edge servers \\
$\mathbb{N}$ & Total incoming resource requests \\
$Q$ & Query in the context of transformer \\
$K$ & Key in the context of transformer \\
$V$ & Value in the context of transformer \\
$W$ & Parameter matrix in the context of transformer \\
$\hat{r}_{\psi}\sim \hat{R}_{\psi}$ & Reward from Math$\Sigma$tral model\\
$\hat{r}_{\mu}\sim \hat{R}_{\mu}$ & Reward from custom model\\
$r_{\tau}$ & Instantaneous reward \\
$\mathbb{R}$ & Reward from individual agent \\
$E$ & Episode \\
$J$ & MASAC objective function \\
\bottomrule
\end{tabular}
\end{table}
\begin{table}[!b]
\caption{System model parameters}
\begin{center}
\begin{tabular}{|c|c|c|c|c|c|}
 \hline
 \textbf{Parameter} & \textit{$B_{f}$} & $\textit{$NF$} $ & \textit{$P_{TX}$} & \textit{$PL$} & \textit{$G_{TX}$} \\
 \hline
 \textbf{Value} & 20 MHz & 7dB & 46 dBm & 125 dB & 5.5 dBi \\
 \hline
\end{tabular}
\label{tableSM}
\end{center}
\end{table}
\subsection{Foundations for reasoning using LLM}
The transformer architecture within LLMs allow them to translate the intent behind the information provided. Long-range dependencies between the sequence of words is established using self-attention mechanism which in turn is used for the reasoning \cite{self-attention}. Three main components are the query $Q$, the key $K$ and the value $V$. Consider a sequence of token embeddings $w = [w_1,w_2,w_3,....,w_n]$ for a sequence of length $n$. The dimensionality of the queries and keys is $d_k$ while that of the values is $d_v$. Additionally, consider the linear projections of the embeddings using learned parameter metrices $W^Q, W^K, W^V$ with respect to queries, keys and values. A softmax layer is applied to scaled dot-product of queries, keys and values produce the attention weights as in (\ref{self-attention dot-product}).
\begin{equation}
    \text{Attention}(Q, K, V) = \text{softmax}\left( \frac{QK^\top}{\sqrt{d_k}} \right) V
    \label{self-attention dot-product}
\end{equation}

\noindent The single attention mechanism is further improved by computing multiple projections $(W_Q^{(i)}, W_K^{(i)}, W_V^{(i)})$ from different sub-spaces to enhance information representation. In transformers, multiple independent $h$ attention functions are used as multi-head attention mechanism as in (\ref{heads}) and (\ref{multi-head attention}).
\begin{equation}
    \text{head}^i = \text{Attention}(Q^i, K^i, V^i)
    \label{heads}
\end{equation}
\begin{equation}
    \text{MultiHead(Q,K,V)} = \text{Concat(head}^1,....,\text{head}^h)\text{W}_O
    \label{multi-head attention}
\end{equation}
\section{RAGAM-AI design}
This section navigates through the particulars of the conducted simulations. The section begins by defining the components of the Markovian Decision Process (MDP) formulation used by the RAGAM-AI algorithm. The experimental setup with the parameters that were utilized is further presented alongside the overall implementation process.    
\subsection{Markovian approach}
The decision-making within this work uses a Markovian approach as most of the RL algorithms do. The system in consideration is memoryless depending solely on the current state. Therefore, the MDP represented as  $\mathcal{(S,A,P,R,\gamma)}$ is elaborated further as follows:
\subsubsection{State}
The state space is a set of different resource allocations and demands alongside the performance metrics latency and data rate. In (\ref{state space}) $a$ and $d$ are the allocations and demands respectively for each agent $n\hspace{0.05cm} \epsilon \hspace{0.05cm}\mathrm{N}$.
\begin{equation}
    \mathcal{S} = \{c_a^n,m_a^n,b_a^n,c_d^n,m_d^n,b_d^n,L^n,D^n\} \hspace{0.1cm} n \epsilon \mathrm{N}
    \label{state space}
\end{equation}
\subsubsection{Actions}
The action space is a set of resource allocations $a_{resource}$ predicted by the SAC policy given the observation from the environment for each agent. 
\begin{equation}
    \mathcal{A} = \{a_{CPU}^n,a_{mem}^n,a_{bw}^n\} \hspace{0.1cm} n \epsilon \mathrm{N}
    \label{action space}
\end{equation}
\subsubsection{Reward function}
\begin{figure}[t]
    \centering

    \begin{subfigure}[b]{0.53\textwidth}
        \centering
        \includegraphics[width=0.7\textwidth]{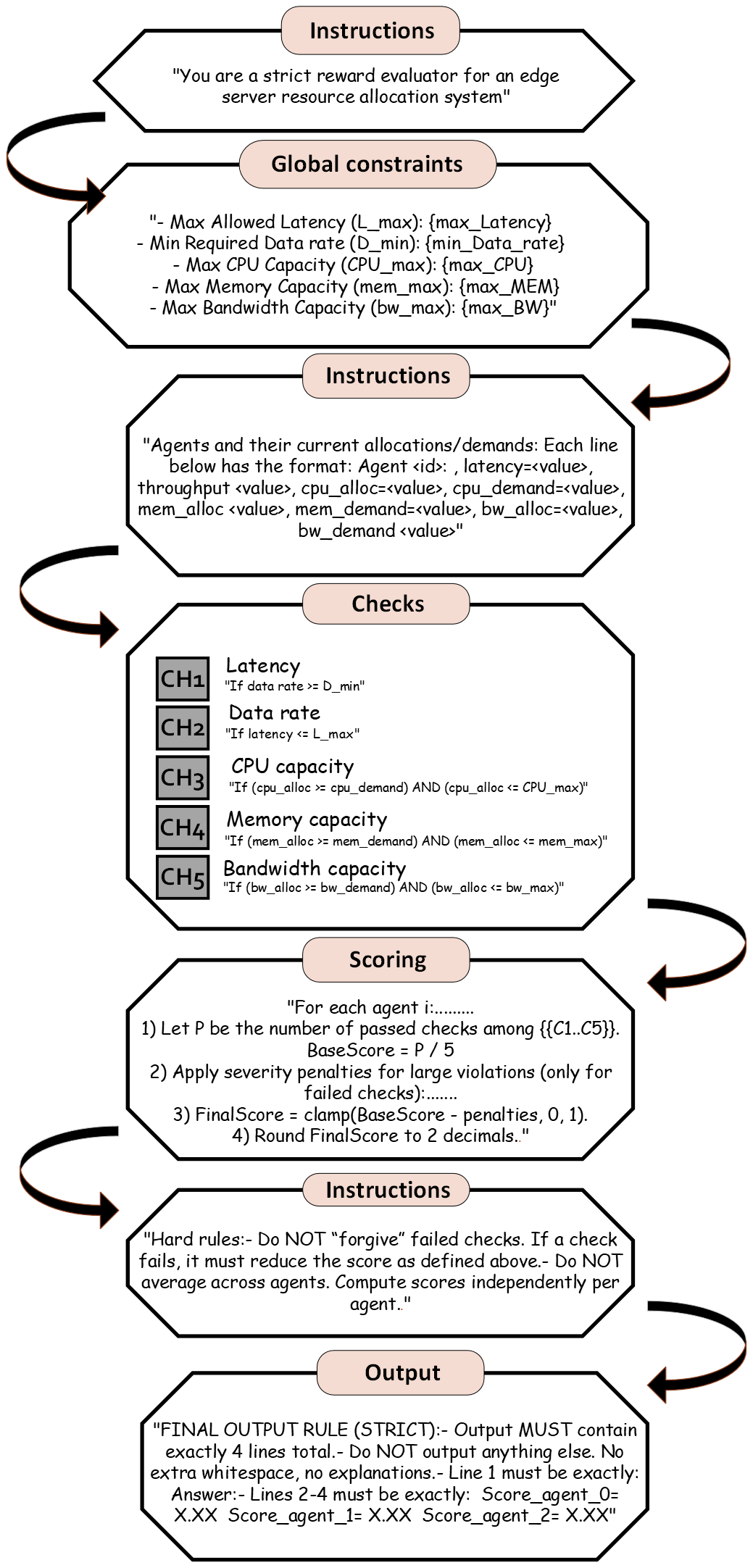}
        \caption{CoT prompt }
        \label{fig:sub1}
    \end{subfigure}
   
    \begin{subfigure}[b]{0.55\textwidth}
        \centering
        \includegraphics[width=0.9\textwidth, trim=1cm 10.5cm 1cm 14cm,
    clip]{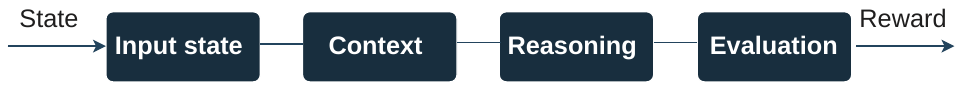}
        \caption{Prompt pipeline}
        \label{fig:sub2}
    \end{subfigure}

    \caption{CoT-based reward generation workflow}
    \label{fig:cot}
\end{figure}

The innovation for this work lies within the reward function of the formulation. The reward is designed using two sub-parts. In the first half to stabilize early training, the reward $r_\tau$ is obtained by alternating between a custom reward function $\hat{R}_{\mu}$ and the Math$\Sigma$tral model $\hat{R}_{\psi}$ as in (\ref{reward_decay}). $\hat{R}_{\psi}$ is developed by querying using states $(s_0,s_1,...,s_t)$ within CoT style prompting as in Fig.\ref{fig:cot} to obtain the instantaneous reward $\hat{r}_{\psi}$. The states provide values which are passed within the prompt instructions in a structured manner as visible in Fig.\ref{fig:cot} for going through series of constraint checks before being evaluated by the Math$\Sigma$tral model. On the other hand, $\hat{r}_{\mu}$ is the instantaneous custom reward sampled from $\hat{R}_{\mu}$ that awards ($w$) or penalizes ($\lambda$) the constraints $f(k)$ based on the satisfaction ($\mathbb{I}_{(\Omega \le 0)}$). The parameters ($w$) and ($\lambda$)  are empirically chosen by heuristics used to regulate the trade-off across multiple constraints while $f(k)$ quantifies the deviation associated with constraint $k$. The sum of the awards and penalties is further adjusted for considering individual contribution $\mathbb{R}$ from each agent as seen in (\ref{custom_reward}). The probability of using $\hat{R}_{\psi}$ or $\hat{R}_{\mu}$ is linked to $p_\psi(e)$, a stepwise decaying function dependent on steps $e$. The decaying function is added to gradually transition from the rewards generated by the LLM model to the surrogate model. During early training episodes, the framework queries the LLM more frequently while decaying exponentially to allow the surrogate model approximation for later. 
\begin{equation}
r_{\tau} =
\begin{cases}
\hat{r}_{\psi}\sim \hat{R}_{\psi}, & p_\psi(e) \\
\hat{r}_{\mu}\sim \hat{R}_{\mu}, & 1 - p_\psi(e)
\end{cases}
\label{reward_decay}
\end{equation}
\begin{equation}
    \hat{R}_{\mu}= \frac{1}{\mathbb{R}}\sum_{i=0}^{T}
        \begin{cases}
        w f(k), & \text{if} \hspace{0.1cm} \mathbb{I}_{(\Omega \le 0)} \\
        \lambda f(k), & \text{otherwise}
        \end{cases}
        \label{custom_reward}
    \end{equation}
After collecting sufficient samples, in the second half a surrogate DNN model is trained to give the reward score $\hat{r}_{\tau}$. The initial supervised dataset is provided by the state-reward pairs from LLM inferencing. Later, a small DNN architecture is trained to approximate the LLM-generated state-reward pairs. Further, the surrogate model is put to use only after it is trained enough to have reasonably low Mean Absolute Error (MAE). In (\ref{mae}), $y_i$ is the true value and $\hat{y}_i$ is the predicted value from the surrogate model trained over $\mathcal{N}$ samples. Additionally, the MAE is subjected to a threshold $\epsilon$ to maintain the average prediction error between the rewards predicted by the different models to be under 10\%.  
\begin{equation}
\text{MAE} = \frac{1}{\mathcal{N}}\sum_{i=1}^{N} \left| y_i - \hat{y}_i \right|
\label{mae}
\end{equation}
\subsection{Simulation setup}
\begin{figure*}[!t]
     \centering
        \includegraphics[trim=7cm 7cm 10cm 2cm, clip,width=1.1\textwidth]{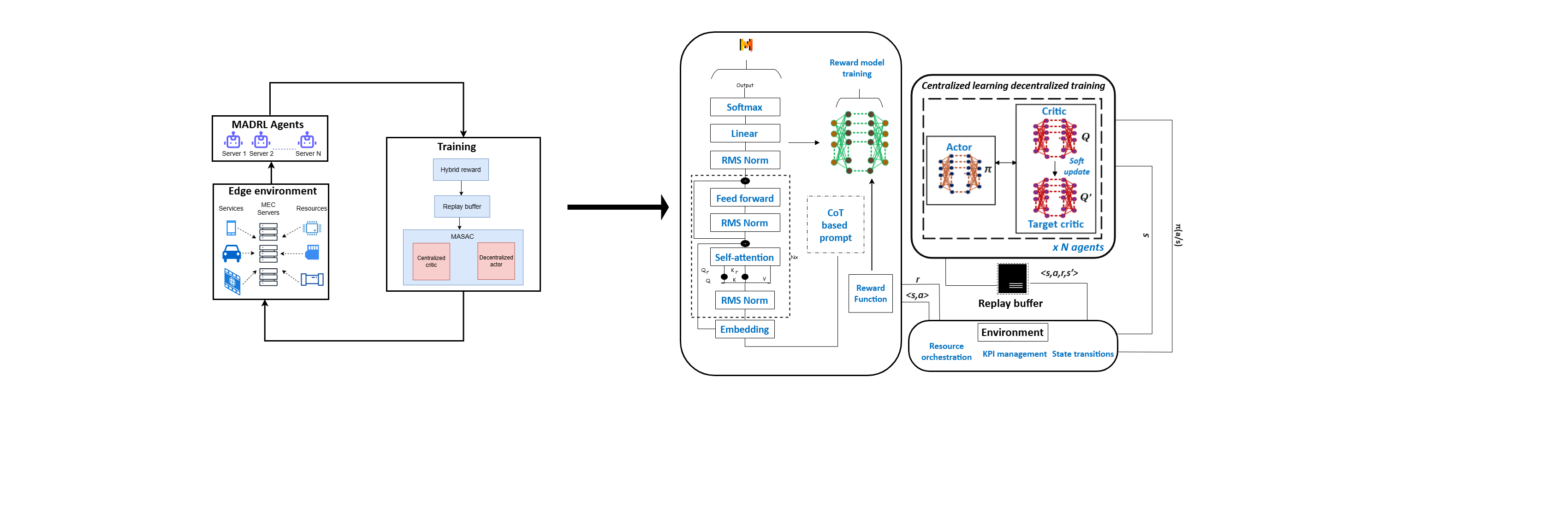}
       \caption{RAGAM-AI algorithm architecture}
       \label{Model}
\end{figure*}

Fig.\ref{Model} depicts the architecture behind the proposed algorithm. A multi-agent actor-critic network is structured within a centralized learning but decentralized execution framework. Further, the architecture includes the reward function utilizing the Math$\Sigma$tral model while training a DNN after enough samples are collected. Algorithm \ref{algorithm1} explains the simulation process step-by-step. The LLM model-based reward function is wrapped inside the MASAC implementation for receiving feedback to build an Agentic-AI solution. Here, the principle of RLAIF as mentioned before is applied. The experimental setup is further explained in detail below.    
\subsubsection{Dataset}
For training purposes, the resource request available from Microsoft Azure Packing traces is used \cite{azure}. The publicly available dataset captures workloads from virtual machines, containers and tasks containing essential information about CPU, memory and bandwidth. These are large-scale data collected under varying realistic operational conditions. The traces were first cleaned by removing incomplete records and entries with missing or invalid resource usage values. A consistent sorting was applied to make sense of the workload ids and timestamps. The necessary resources such as CPU, memory and Bandwidth utilization value was retrieved. Hence, for this work the packing traces after extensive pre-processing were used as the resource requests. 
\subsubsection{Training conditions}
A MASAC policy is used for the resource management at the edge in this work. The training is further augmented by bringing Gen-AI components as discussed on multiple occasions. The values for all the parameters that were used are mentioned in Table \ref{tab-param}. Each agent iterates over a set of episodes at the end of which the training weights are updated and the RL environment is reset as visible from algorithm \ref{algorithm1}. The objective $J(\pi)$ of the policy is to maximize the long-term cumulative reward $r_t$ while enforcing entropy regularization $\mathcal{H}(\pi_i(\cdot\mid o_{i,t}))$ as in (\ref{MASAC-eq}). 
\begin{equation}
    J(\pi)=
\mathbb{E}_{\tau\sim\rho_\pi}
\left[
\sum_{t=0}^{\infty}\gamma^t
\left(
r_t+\sum_{i=1}^{N}\alpha_i\,\mathcal{H}(\pi_i(\cdot\mid o_{i,t}))
\right)
\right]
\label{MASAC-eq}
\end{equation}
\subsubsection{Language model settings}
As explained previously, the reward function utilizes the Math$\Sigma$tral model for generating rewards using CoT prompting. These design choices are made strategically keeping in mind the requirements of the problem formulated in section III. Math$\Sigma$tral is a specialized model for mathematical reasoning. It is fine-tuned specifically on mathematical data to generate reliable mathematical outputs with detailed explanatory solutions \cite{mathstral}. However, for this model to successfully generate the desired outputs, the prompts passed on to the model have to be immaculate. Therefore, given the sequential decision-making process of the problem in hand, a CoT based prompt is designed as in Fig.\ref{fig:cot}. A step-by-step approach is more structured and thus avoids reasoning errors by validating each constraint in comparison to any other prompting styles. Additionally, the prompt is not user-driven but rather a fixed, system-controlled component of the RAGAM-AI pipeline.

\begin{table}[b]
\caption{Experimental Parameters}
\begin{center}
\begin{tabular}{|c|c|c|}
 \hline
  \textbf{Setting} & \textbf{Parameter} & \textbf{Value}\\
 \hline
 
 \multirow{4}{*}{\textbf{MASAC}} & \textit{Actor learning rate} & 0.001\\
                        & \textit{Critic learning rate} & 0.0001\\ 
                        & \textit{Hidden layers} & 8\\
                        & \textit{Neurons per layer} & 64\\

\hline
\multirow{2}{*}{\textbf{Math$\Sigma$tral}}&\textit{Temperature} & 0.2\\
                        & \textit{Token} & 60\\ 
\hline
\multirow{3}{*}{\textbf{GPU}}&\textit{Model} & NVIDIA 6000\\
                        & \textit{CUDA Version} & 12.6\\ 
                        & \textit{Memory} & 49140MiB\\

\hline
\multirow{8}{*}{\textbf{Simulation}} & \textit{Episodes} & 25\\
                                  & \textit{Steps} & 512\\
                                 & \textit{Optimizer} & Adam\\
                                 & \textit{Activation} function & ReLu\\
                                 & \textit{Discount factor ($\gamma$)} & 0.90\\
                                 & \textit{Batch size} & 64\\
                                 & \textit{Number of DNN layers} & 8\\
                                 & \textit{Layer size} & 64\\
                                 & \textit{Temperature parameter ($\alpha$)} & 0.5\\ 
\hline
\multirow{3}{*}{\textbf{Implementation}}&\textit{Language} & Python 3.11.14\\
                        & \textit{RL Library} & Tensorflow 2.20.0\\ 
                        & \textit{LLM Library}& PyTorch 2.9.1\\
\hline
\end{tabular}
\label{tab-param}
\end{center}
\end{table}

\begin{algorithm}[!htbp]
\SetKwFunction{Step}{Step}
\SetKwProg{Fn}{Function}{:}{}
\caption{RAGAM-AI}
\label{algorithm1}
\For{agent \(i=1,\dots,M\)}{
    Initialize parameters: actor \(\theta_i\), critic \(\phi_i\), temperature \(\alpha_i\), target critic \(\phi'_i \gets \phi_i\) \\ 
    Initialize replay buffer \(\mathcal{D}\)
}
\For{episode =1 to N}{
    environment ← environment.reset() \\
    state ← initial state \(s\)\\
        \For{steps \(=1\) to \(P\)}{
        Collect observation \(o_i\)\\
        Sample action for each agent \(a_i \sim \pi_{\theta_i}(\cdot\mid o_i)\)\\
        \Fn{\Step{$s,a$}}{
            \Fn{{$R(s,a)$}}{
            \If{step mod K}{
            train reward model}
            \If{MAE $<$ threshold $\epsilon$}{
            r ← Score using reward model\\
            }
            \Else{
            Sample $u \sim \mathcal{U}(0,1)$\\
            \If{$u <p_\psi(e)$}{r ← $\hat{R}_\mu$}
            \Else{Build CoT prompt based on $s$\\
            r ← $\hat{R}_\psi$\\}
            Store in reward cache
            }
            \Return r\\}
            Collect next state $s'$\\
            Collect done signal\\
            \Return $(s', r, done)$\\
            }
        Store transition \((s,a,r,s')\) in \(\mathcal{D}\)\
            \(s \gets s'\)\\\
            \If{training step}{
                Sample a minibatch from \(\mathcal{D}\)\ for all agents\\
                Update parameters \(\theta_i\), \(\phi_i\), \(\alpha_i\), \(\phi_i'\)
        }
         $s$ ← $s'$\\
         Log values
    }
    Store model weights
}
\end{algorithm}
\section{Results}
\begin{figure}[t]
    \centering
    \includegraphics[width=1.0\linewidth]{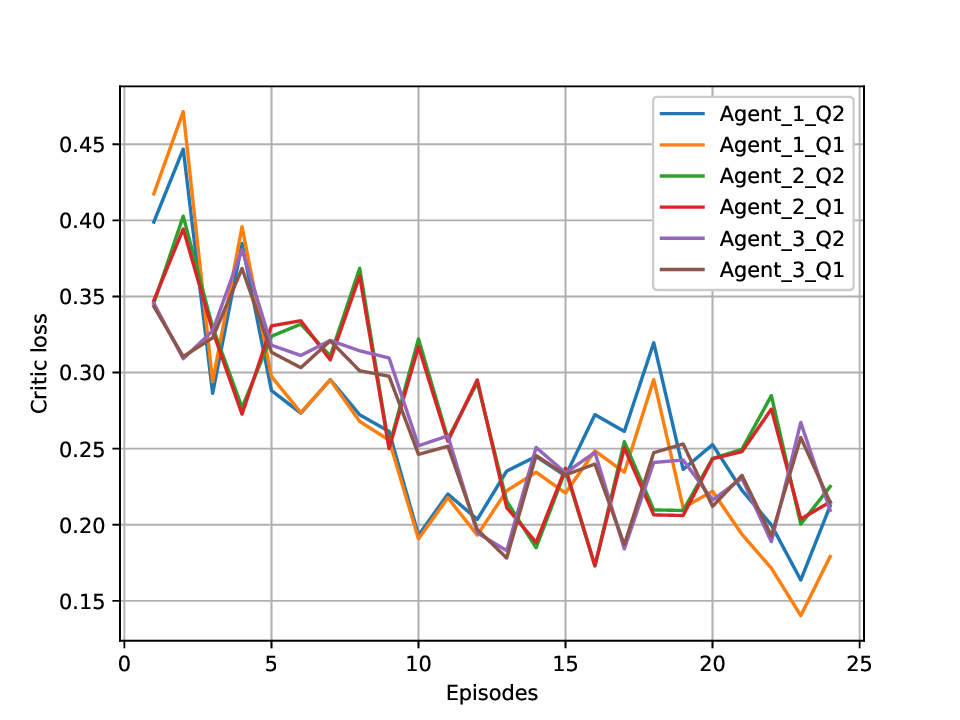}
    \caption{Training loss}
    \label{fig:training_loss}
\end{figure}

This section discusses the baseline policies used for comparison against the performance of RAGAM-AI. The results are further elaborated in terms of resource allocation, performance costs and the capacity to adapt to scaling networks. For each plot, the results from the steps of each episode is averaged to observe performance over the training episodes. Further, the critic loss convergence from the two critic networks within the three considered RL agents is shown in Fig.\ref{fig:training_loss}. Also, to provide evidence for reward stability on using LLM, fifty random states are subjected to multiple LLM calls and three of those observations are reported in Table \ref{tab:reward_stability}. It was concluded that rewards were stable even after multiple runs and hence the LLM-based reward design is able to consistently output the correct rewards.  
\begin{table}[t]
\centering
\caption{Reward stability study}
\begin{tabular}{cccc}
\hline
\multicolumn{4}{c}{\textbf{Data}}\\
\hline
\multicolumn{2}{c}{\textit{Input States}} &
\multicolumn{2}{c}{50} \\
\hline

\multicolumn{2}{c}{\textit{Query runs}} &
\multicolumn{2}{c}{5} \\
\hline
\multicolumn{4}{c}{\textbf{Reward stability}}\\
\hline
\textit{} & \textit{Mean} & \textit{Standard deviation}& \textit{Confidence}  \\
\hline
\textit{State 1} & 0.88 & 0.06 & $\pm$0.07 \\
\hline
\textit{State 2} & 0.68 & 0.08 & $\pm$0.1 \\
\hline
\textit{State 3} & 0.73 & 0.06 & $\pm$0.07 \\
\hline
\end{tabular}
\label{tab:reward_stability}
\end{table}

\subsection{Baselines}
The performance of the RAGAM-AI Algorithm is compared with a range of baselines with different complexities. All baselines are evaluated under the same experimental conditions using the same dataset, parameters and evaluation metrics to ensure a fair comparison. The following methods were implemented for benchmarking: 
\begin{itemize}
    \item \textit{SAC:} A single agent SAC algorithm with a traditional reward function (\ref{custom_reward}) is used to highlight the performance without the support of Gen-AI components. 
    \item \textit{REINFORCE:} The REINFORCE approach optimizes the policy parameters without learning the value function. It utilizes Monte-Carlo sampling for computing the estimates of policy gradients as in (\ref{reinforce}) while increasing the probability of actions that yield higher returns \cite{williams}. This approach is included to highlight the performance of a fairly less complex model-free policy in terms of learning stability while compared to an advanced RL algorithm. 
    \begin{equation}
    \nabla_{\theta} J(\theta)
    =
    \mathbb{E}_{\pi_\theta}
    \left[
    \nabla_{\theta} \log \pi_\theta(a_t \mid s_t) \, G_t
    \right]
    \label{reinforce}
    \end{equation}
    
    \item \textit{Random:} In this approach, resources are assigned using uniform random sampling with respect to the constraints. A plausible set of allocations are selected without exceeding the server capacities.
\end{itemize}
\subsection{Resource allocation}
\begin{figure}[!t]
\centering
\begin{subfigure}[b]{0.5\textwidth}
    \centering
    \includegraphics[width=\textwidth]{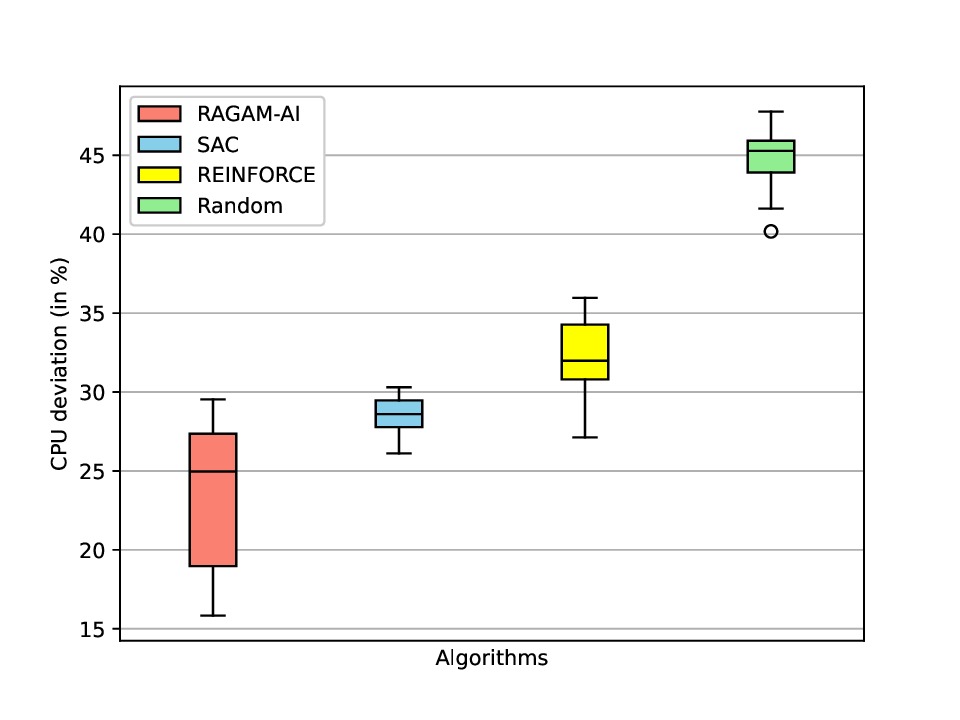}
    \caption{CPU deviation}
\end{subfigure}
\hfill
\begin{subfigure}[b]{0.5\textwidth}
    \centering
    \includegraphics[width=\textwidth]{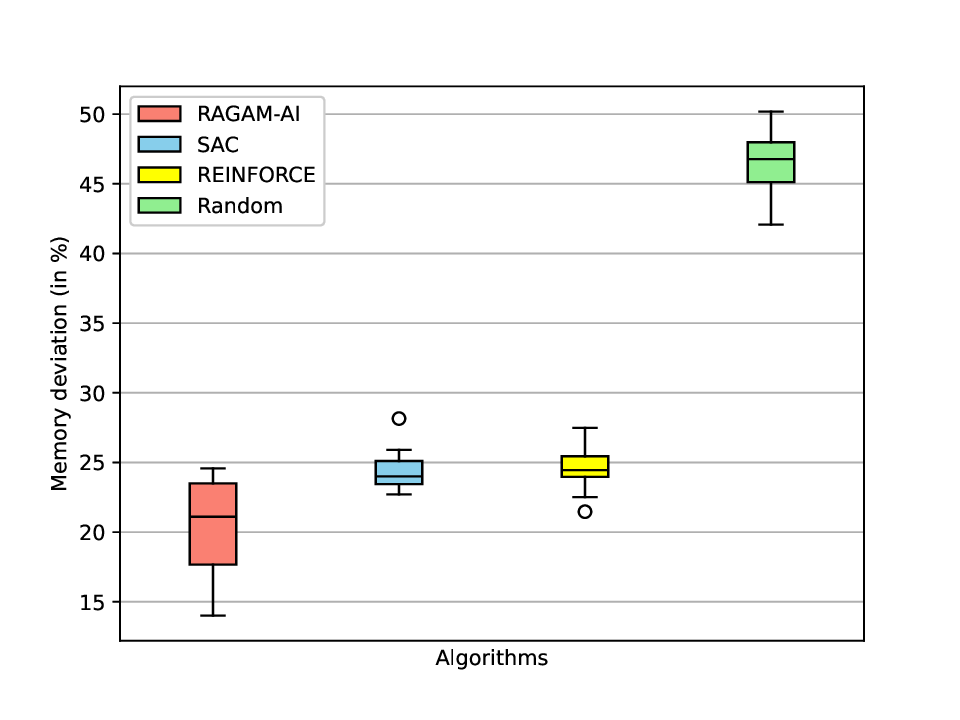}
    \caption{Memory deviation}
\end{subfigure}
\hfill
\begin{subfigure}[b]{0.5\textwidth}
    \centering
    \includegraphics[width=\textwidth]{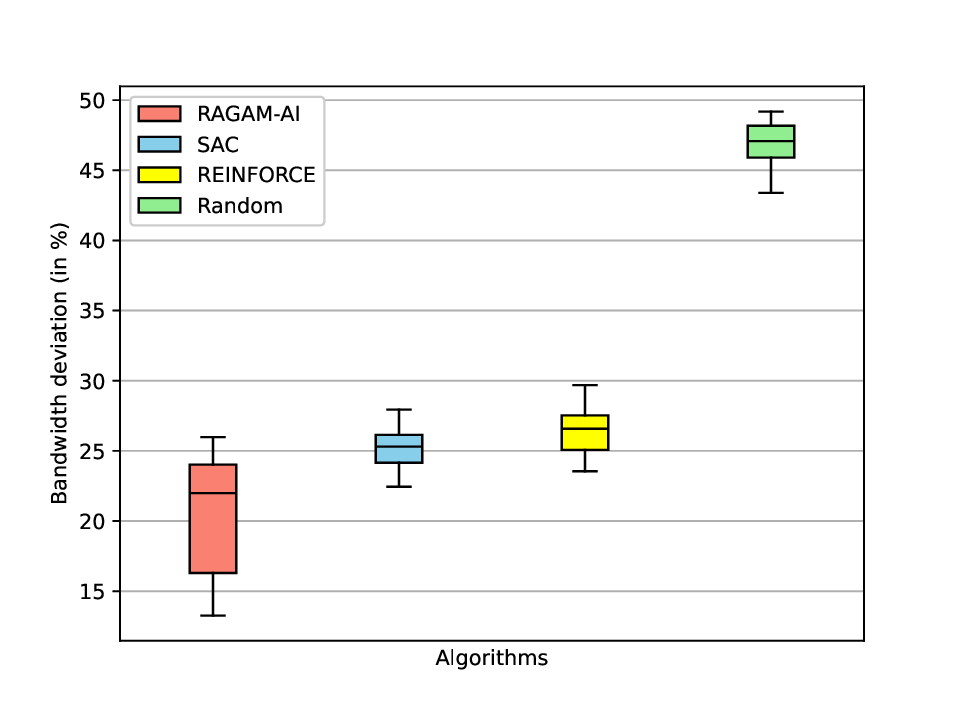}
    \caption{Bandwidth deviation}
\end{subfigure}
\caption{Resource allocation performance}
\label{Resource allocation}
\end{figure}

Resource allocation by RAGAM-AI is evaluated by measuring the difference between the demand $d$ and allocations $a$ normalized to the total capacity. Fig.\ref{Resource allocation} showcases the deviations achieved while using CPU, memory and bandwidth resources. The proposed RAGAM-AI exhibits improved performance in resource allocation by consistently reducing the deviation between the allocated resources and the observed demand. There is a consistent performance improvement by RAGAM-AI relative to baseline policies SAC, REINFORCE and random as reported in Table \ref{tab:improvement}. The baseline policies achieve higher deviations and costs leading to poor performance. While Table \ref{tab:improvement} provides the gain on comparing the mean values of the deviations, Fig.\ref{Resource allocation} provides a visual representation of the overall deviation range. The results ascertain the fact that RAGAM-AI is able to take advantage of the hybrid nature of the reward function by successfully guiding the MASAC policy towards making efficient resource allocation decisions. Further, it can be established that using the occasional feedback from the LLM has introduced contextual reasoning enhancing the reward signals. The entropy-regulated SAC architecture when combined with LLM-assisted reasoning reliably maps resource supply and system demand enabling in a robust performance by RAGAM-AI.   
\begin{table}[t]
\centering
\caption{Performance gain relative to baselines}
\begin{tabular}{lccc}
\hline
\textbf{Metric} & \textbf{SAC} & \textbf{REINFORCE} & \textbf{Random} \\
\hline
\textit{CPU deviation} & 17.33\% & 27.29\% & 47.36\% \\
\hline
\textit{Memory deviation} & 16.38\% & 17.50\% & 56.09\% \\
\hline
\textit{Bandwidth deviation} & 19.27\% & 23.01\% & 56.52\% \\
\hline
\textit{Latency cost} & 2e-4 & 2.4e-4 & 1.42e-3 \\
\hline
\textit{Data rate cost} & 2e+6 & 3e+6 & 7e+6 \\
\hline
\end{tabular}
\label{tab:improvement}
\end{table}

\subsection{Performance cost}
\begin{figure}[!t]
\centering
\begin{subfigure}[b]{0.44\textwidth}
    \centering
    \includegraphics[width=\textwidth]{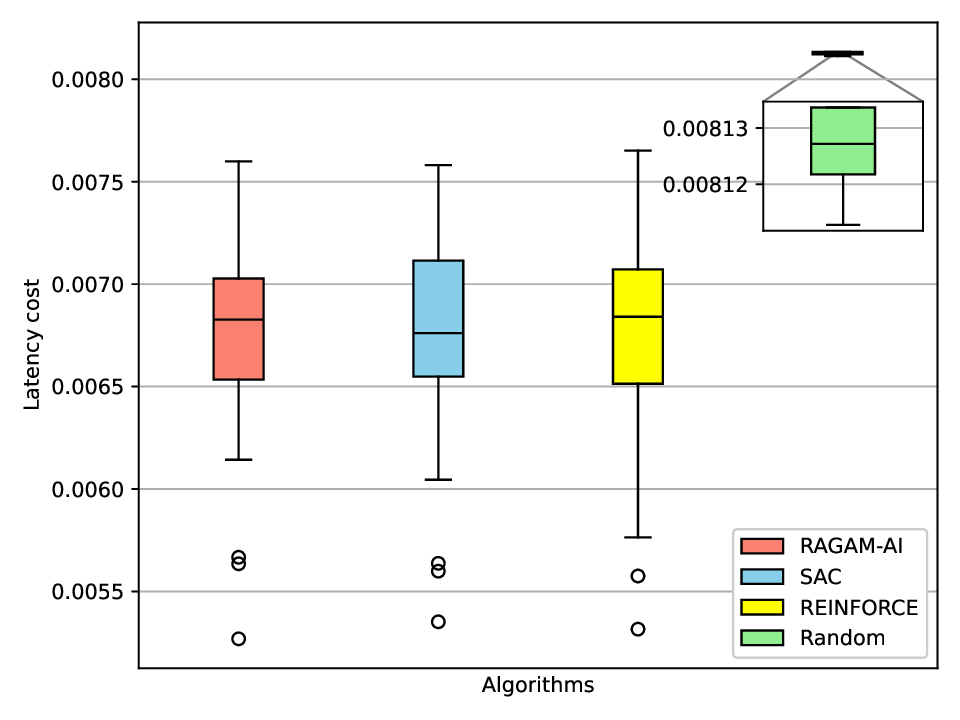}
    \caption{Latency cost}
\end{subfigure}
\hfill
\begin{subfigure}[b]{0.44\textwidth}
    \centering
    \includegraphics[width=\textwidth]{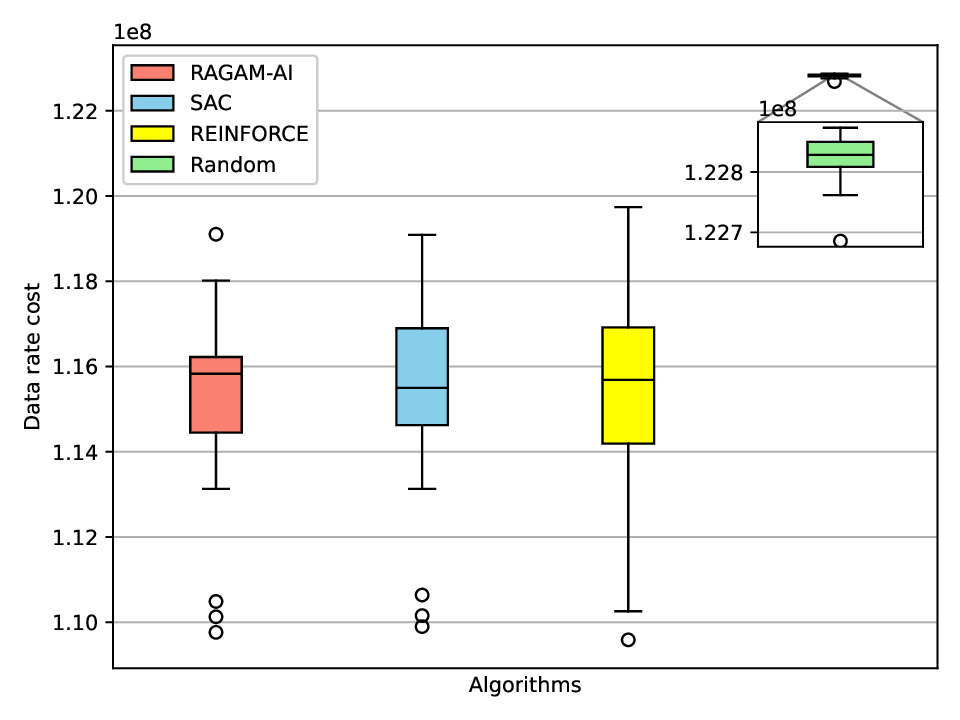}
    \caption{Data rate cost}
\end{subfigure}

\caption{Performance costs}
\label{performance_cost}
\end{figure}

In addition to demand-allocation deviation, performance cost is evaluated in terms of latency and data rate cost. These metrics are weighed in by utilizing the allocated and total bandwidth to include budget-aware efficient resource allocations. Further, the performance metrics should be evaluated relative to the amount of resources consumed and prevents bias toward larger allocations. It can be visualized in Fig.\ref{performance_cost} that RAGAM-AI yields the least performance cost in terms of both latency and data rate in comparison to the considered baselines. Similar to the resource allocation, a cost improvement can be observed from Table \ref{tab:improvement}.  Enhanced reward shaping utilizing the exploration from the MASAC policy and reasoning capabilities of Math$\Sigma$tral capture the multi-objective constraints of the system better than the other algorithms. Further, RL optimizes long-term decision policies through interaction, while the LLM provides contextual priors leading to more efficient resource utilization under budget constraints.  
\subsection{Adaptability}
In this work, the multi-agent RAGAM-AI based on RL-LLM framework with a systematic reward mechanism contributes significantly to system scalability and adaptability. The decentralized learning component within the RL algorithm enables each individual agent to learn policies that optimize local decision-making while adhering overall system objectives designed within the reward function. It is to be noted that the reward is not directly optimizing the resource deviations or KPI costs, rather it allows the framework to learn the network demand patterns while keeping the constraints in check. With the assistance from LLM, agents are able to interpret contextual information and introduce feedback using its pre-trained knowledge. The hybrid approach within the reward function further helps in the development of a surrogate reward model that acts a substitute replacing the otherwise undetermined task. Although the proposed RAGAM-AI method exhibits a larger spread in Fig. \ref{Resource allocation} and Fig. \ref{performance_cost}, it can be inferred as a consequence of enhanced exploration and variability in the learned policy during training. Despite this variability, the RAGAM-AI method consistently achieves lower median cost, indicating superior performance compared to baseline approaches. The agentic system thus built iteratively adjusts its actions dynamically to meet the heterogeneous resource demands to maximize the performance. Overall, the system complexity is addressed with respect to two components: the MADRL setup and the LLM-based reward mechanism. Neither of the components require continuous inferencing on resource-constrained devices. Further, the surrogate reward model alleviates the complexity issues by enabling lightweight decision-making. Therefore, the proposed RAGAM-AI framework is more suitable for edge controllers or orchestration nodes in terms of complexity. 
\section{Conclusions}
To summarize, in this work a hybrid Agentic-AI framework RAGAM-AI is implemented by combining the abilities of RL and Gen-AI. The multi-agent system was developed using the MASAC policy while receiving feedbacks from the Math$\Sigma$stral inside its reward function. Network resources such as CPU, memory and bandwidth are managed at the network edge using RAGAM-AI. The main objective was to strengthen the performance of the MADRL system by augmenting it with context-aware reasoning. Hence, RAGAM-AI demonstrates effectiveness of RLAIF concept by obtaining smaller resource deviations and performance costs in comparison to the baselines. For future work, a workflow encompassing different agents conveying intents from different components such as Open-Radio Access Network (O-RAN) and cloud can be integrated. Further, this work serves as simulation for the proof-of-concept idea proposed and thus can be tested by deploying in cloud-native environments as well later. Broader metrics and processes such as caching may also be considered. Including system-level analysis in terms of runtime overhead measurements, convergence analysis, and sensitivity studies can also provide important additional insights for reducing the system complexity.    



\newpage

 




\vfill

\end{document}